\documentclass[letterpaper]{article} 
\usepackage[arxiv]{aaai24}  
\usepackage{times}  
\usepackage{helvet}  
\usepackage{courier}  
\usepackage[hyphens]{url}  
\usepackage{graphicx} 
\usepackage{natbib}  
\usepackage{caption} 
\usepackage{algorithm}
\usepackage{algorithmic}

\usepackage{newfloat}
\usepackage{listings}
\DeclareCaptionStyle{ruled}{labelfont=normalfont,labelsep=colon,strut=off} 
\floatstyle{ruled}
\newfloat{listing}{tb}{lst}{}
\floatname{listing}{Listing}
\usepackage{subcaption}
\usepackage{pgfplots, pgfplotstable}
\pgfplotsset{compat=1.18}
\pgfplotsset{
    select row/.style={
        x filter/.code={\ifnum\coordindex=#1\else\fi}
    }
}

\title{AI-assisted Gaze Detection for Proctoring Online Exams}
\author{
    Yong-Siang Shih\textsuperscript{\rm 1},
    Zach Zhao\textsuperscript{\rm 1},
    Chenhao Niu\textsuperscript{\rm 1},
    Bruce Iberg\textsuperscript{\rm 2},
    James Sharpnack\textsuperscript{\rm 1}, \\
    Mirza Basim Baig\textsuperscript{\rm 1}
}
\affiliations{
    Duolingo, Inc. \\
    \textsuperscript{\rm 1}\{yongsiang,zach,chenhao,james.sharpnack,basim\}@duolingo.com, \textsuperscript{\rm 2}bruceiberg@duolingocontractors.com
}

\usepackage{amsmath}
\usepackage{amsfonts}

\newcommand{\calX}{\mathcal{X}}
\newcommand{\calA}{\mathcal{A}}

\newcommand{\vXi}{\mathbf{X}^{(i)}}

\newcommand{\va}{\mathbf{a}}
\newcommand{\vai}{\mathbf{a}^{(i)}}

\begin{document}

\maketitle

\begin{abstract}

For high-stakes online exams, it is important to detect potential rule violations to ensure the security of the test. In this study, we investigate the task of detecting whether test takers are looking away from the screen, as such behavior could be an indication that the test taker is consulting external resources. For asynchronous proctoring, the exam videos are recorded and reviewed by the proctors. However, when the length of the exam is long, it could be tedious for proctors to watch entire exam videos to determine the exact moments when test takers look away. We present an AI-assisted gaze detection system, which allows proctors to navigate between different video frames and discover video frames where the test taker is looking in similar directions. The system enables proctors to work more effectively to identify suspicious moments in videos. An evaluation framework is proposed to evaluate the system against human-only and ML-only proctoring, and a user study is conducted to gather feedback from proctors, aiming to demonstrate the effectiveness of the system.

\end{abstract}

\section{Introduction}

The adoption of online proctoring systems has grown in recent years~\cite{nigam2021systematic}. Online tests offer greater flexibility because test takers can take the test remotely without going to a specific test center. However, the problem of cheating is a threat to the validity of the test results~\cite{bilen2021online}. Therefore, security measures need to be built to detect and prevent cheating behaviors.

Online proctoring comes in various forms, including live synchronous proctoring where the proctor watches the test taker remotely during the test session, and asynchronous proctoring where video recordings of the test sessions are recorded and reviewed by proctors. In this study, we focus on the application of online proctoring in the Duolingo English Test (DET)~\cite{cardwell2024duolingo}, which is an online, high-stakes English assessment test where a test taker's video is recorded with the test taker's webcam. The test taker's video, the screen recording, the responses, and other relevant information are collected, and proctors review each test session asynchronously.

 We focus our study on the task of detecting if test takers are looking away from the screen suspiciously, as such a behavior could indicate that test takers are consulting external resources. There are two challenges proctors face when examining such behaviors. Firstly, when the exam video length is long, it could be tedious for the proctors to watch the entire video to find all moments where the test taker is looking away. In addition, as pointed out by \citeauthor{belzak2024measuring}~\shortcite{belzak2024measuring}, a test taker could also naturally look at arbitrary spots as part of their cognitive processing. With the limited amount of information available in the exam videos, different proctors' decisions could be less consistent for this task compared to other tasks such as detecting plagiarism.

\begin{figure}[t]
\centering
\includegraphics[width=1\columnwidth]{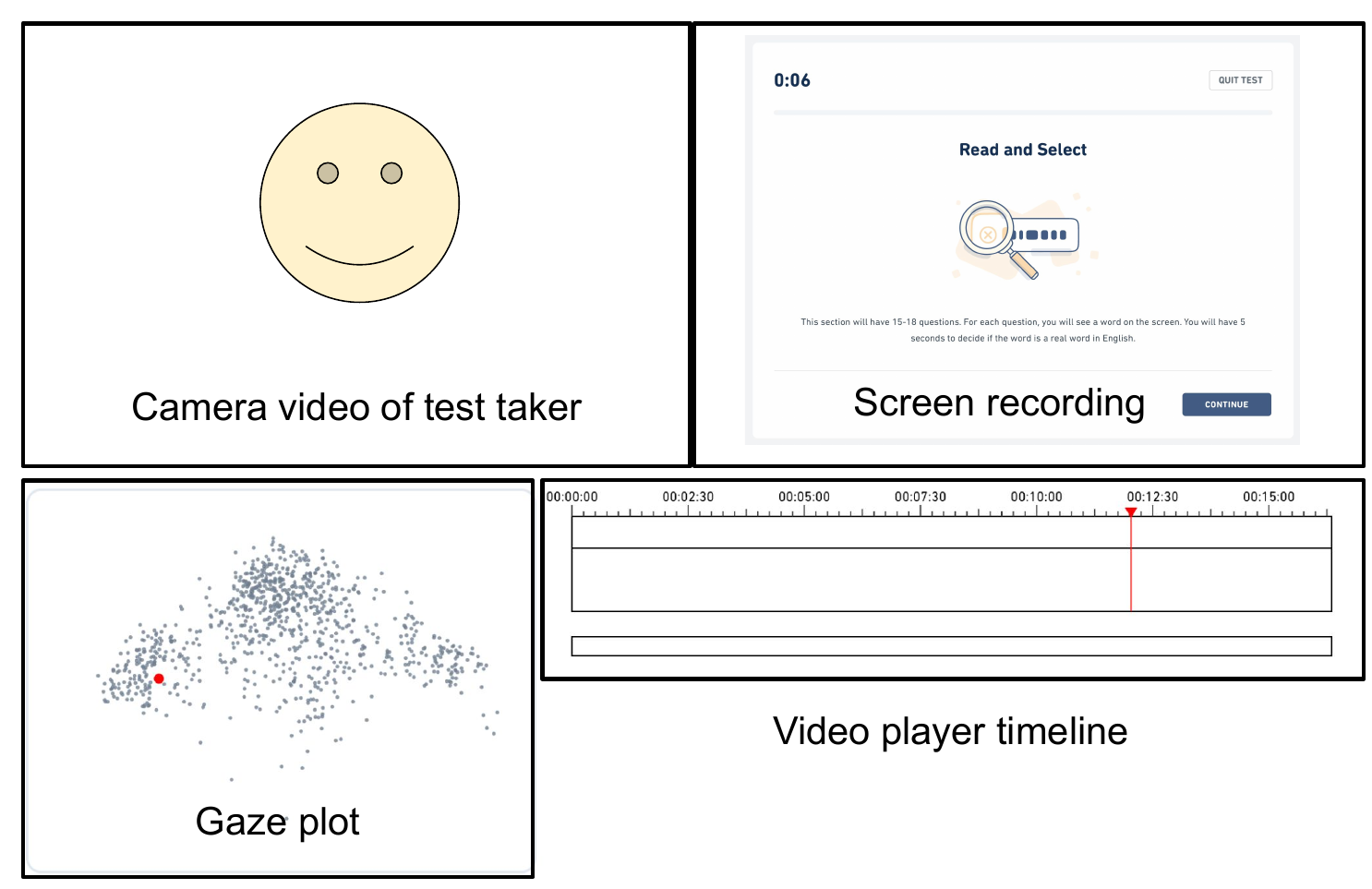}
\caption{The user interface of the gaze system allows proctors to navigate to different video frames using the video player timeline. The points on the gaze plot represent the gaze direction of each frame. When proctors select regions on the gaze plot, the corresponding frames on the timeline would be highlighted.  The gaze direction of the current frame is colored in red in the gaze plot.}
\label{fig.ui}
\end{figure}

In this study, we present an AI-assisted gaze detection system, where the predicted gaze direction of the test taker in each frame is shown on a scatter plot. The user interface is shown in Figure~\ref{fig.ui}. Proctors can select regions on the gaze plot, and the related timestamps on the video player timeline will be highlighted. This allows proctors to navigate to relevant video frames more efficiently, which improves the proctoring experience. In addition, because the system enables a consistent view on the test taker's gaze directions, the quality of the proctoring could also be improved.

We propose to evaluate our AI-assisted gaze detection system against (1) the human-only system, and (2) the ML-only system in an end-to-end fashion. We choose to evaluate the end-to-end performance so that we could capture the effects of the biases that arise when proctors interact with the AI system~\cite{cummings2017automation,selten2023just, bashkirova2024confirmation}. Our framework allows us to properly determine if the proposed system could have a positive impact when deployed into production.

The remaining part of this paper is organized into four sections. Firstly, we describe how the proposed gaze detection system works, including the user interface and how proctors would interact with the system. Secondly, we describe our proposed evaluation framework, including a concrete definition of the task being evaluated. Thirdly, we present the results of a user study where we let proctors try out the system. Finally, we conclude the paper with a discussion on the limitations and future works for our study.

\section{System Overview}

Our system is designed for asynchronous proctoring of online exams. When a test taker takes a test, a video is recorded for the entire test session, and once the video is uploaded, a gaze detection model can be run on each frame of the video to predict the gaze direction in each frame. In practice, a suitable frame rate would need to be selected for inference according to the resource constraints.

\begin{figure}[t]
\centering
\includegraphics[width=0.4\columnwidth]{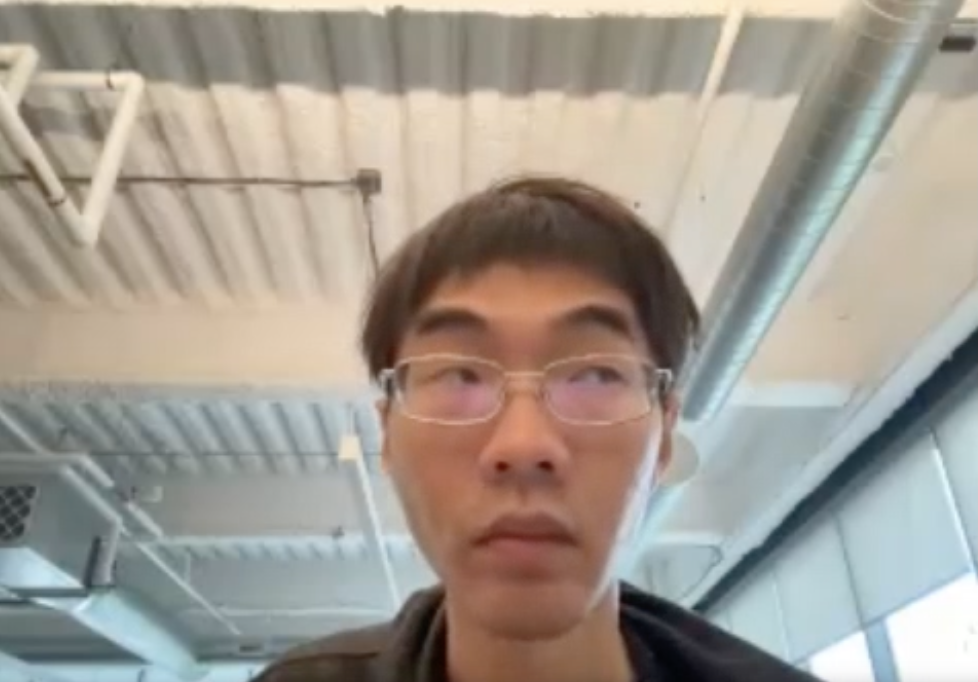}
\includegraphics[width=0.38\columnwidth]{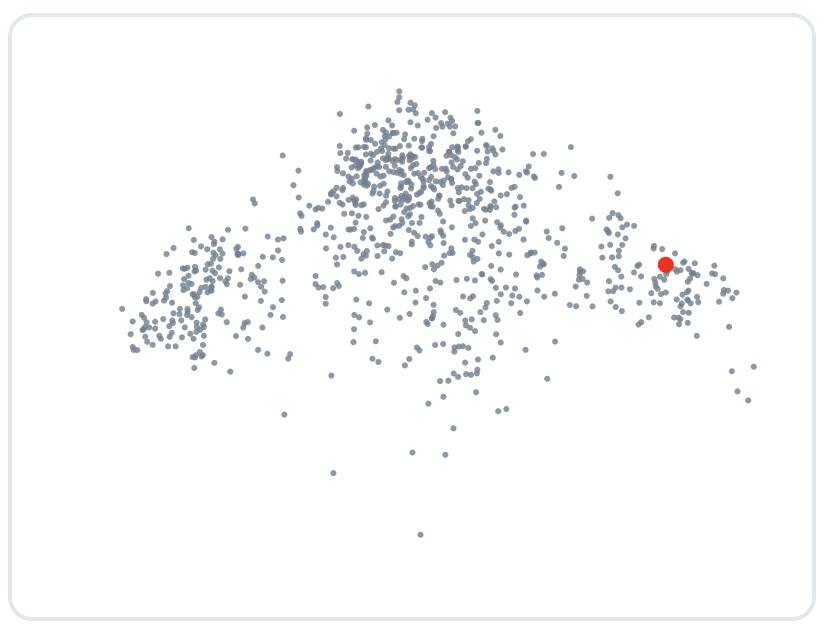}
\caption{Each frame of the test session is shown as a point in the gaze plot, and the position of each point represents the gaze direction in each frame. The current frame's location is colored in red.}
\label{fig.sys}
\end{figure}

The gaze direction predictions will be displayed to proctors as a scatter plot as shown in Figure~\ref{fig.sys}. In particular, the gaze angles predicted by the model can be represented as unit directional vectors originating from the origin, and these vectors are projected onto the 2D plane, with each point on the plot representing a frame and its associated gaze direction. Currently, our gaze plot only represents the gaze directions (i.e., the angles of the gazes), and not the exact location on the screen where the test taker is looking at. However, a similar plot can also be used for models that predict the exact screen location.

Proctors can select regions on the eye gaze plot and the corresponding frames will be highlighted on the video player timeline. This allows proctors to navigate to frames with similar gaze directions within the selected region. For instance, if a proctor observes a suspicious moment when the test taker is looking away from the screen, the proctor can consult the gaze plot to find the current video frame's location, and select a region around it. The system will highlight all other relevant timestamps, allowing the proctor to navigate to those timestamps and confirm whether the test taker is also exhibiting suspicious behaviors at those moments.

\begin{figure}[t]
\centering
\begin{subfigure}[b]{0.45\textwidth}
\centering
\includegraphics[width=0.60\columnwidth]{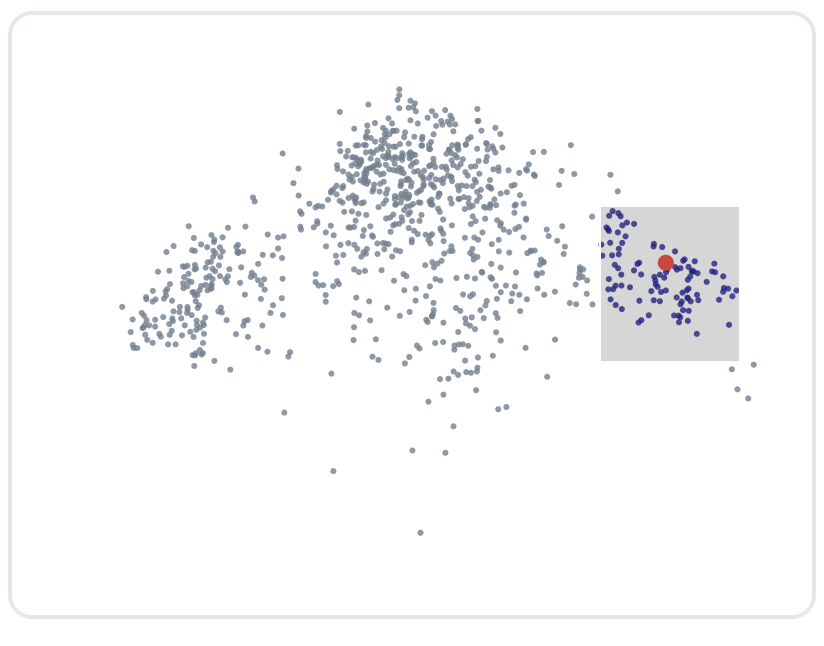}
\caption{The gaze plot with a selected region.}
    \label{timeline_subfig:1}
\end{subfigure}
\hfill
\begin{subfigure}[b]{0.45\textwidth}
\centering
\includegraphics[width=0.90\columnwidth]{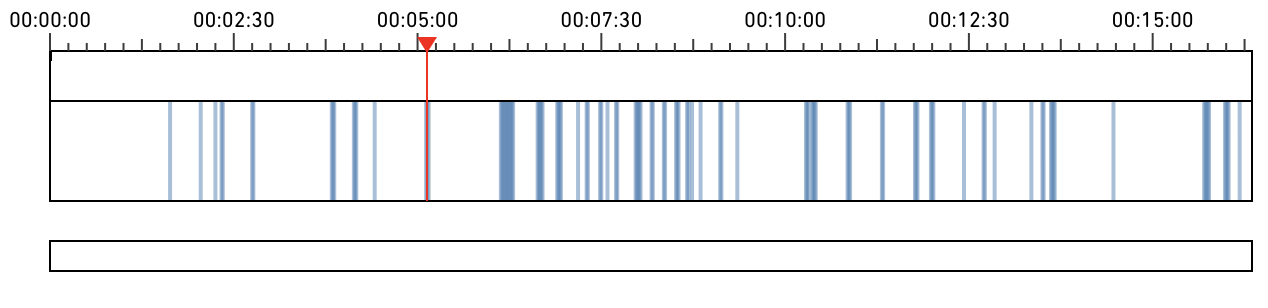}
\caption{The video player timeline with frames highligted.}
    \label{timeline_subfig:2}
\end{subfigure}
\caption{The video player timeline allows proctors to navigate to specific moments by clicking on the desired timestamp. Timestamps where gaze predictions fall within the selected region of the gaze plot are highlighted in blue. The space on top of the timeline can be used to show other notable events. The white bar below is used to select a specific time interval to zoom into.}
\label{fig.timeline}
\end{figure}

\section{Evaluation Framework}

To evaluate the effectiveness of the system in human-based asynchronous proctoring, we apply the concept of \textit{human-ML complementarity} \cite{rastogi2023taxonomy} to define the evaluation goals and propose our experiment plans.

\subsection{Human-ML Complementarity}

In a hybrid decision-making system like AI-assisted proctoring, human-ML complementarity is the condition where the hybrid system outperforms both humans and ML models. Following the notation used by \citeauthor{rastogi2023taxonomy}~\shortcite{rastogi2023taxonomy}, denote $\calX$ as the set of all available features of a given test session, including video recording, responses, scores, etc. Denote the action space as $\calA$, where for a test session with $T$ frames, $\mathbf{a} \in \calA$ is a binary sequence with length $T$, and $\mathbf{a}_t$ indicates whether the $t$ -th frame is labeled positive (i.e. looking away from the screen) or not. Then a decision-making system for labeling gaze direction in a test session can be written as a mapping $\pi: \calX \rightarrow \calA$. Denote $\Pi$ as the set of all possible $\pi$.

In this work, there are three systems of interest: (1) the human-only system $\pi_H$, where human proctors label the test session mainly by watching the video; (2) the ML-only system $\pi_M$, where binary predictions are made by thresholding predicted gaze directions on each frame; and (3) the hybrid system $\pi_{H+M}$, where human proctors label the test session with additional access to predicted gaze directions.

Empirically, it is less likely that a $\pi_M$ is better than $\pi_H$, as the gaze detection system only has access to a frame to make each prediction, while human proctors have more context from the whole test session than a frame. However, it is possible that $\pi_{H+M}$ is a better system than $\pi_H$ and $\pi_M$ through human-ML complementarity. That is, with an evaluation function $F: \Pi \rightarrow \mathbb{R}$, we want to verify human-ML complementarity: $F(\pi_{H+M}) > \text{max} \{F(\pi_{H}),F(\pi_{M})\}$.

\subsection{Proposed Experiments}

On a dataset with $N$ test sessions, we define the evaluation function $F$ as:
\[
F(\pi) = \frac{1}{N} \sum_{i=1}^N s(\vXi, \pi(\vXi))
\]
Where $s: \calX \times \calA \rightarrow \mathbb{R}$ is a scoring function of a labeling result for a given test session, regardless of where the labeling result comes from. Without access to ground truth labels, we use a labeling process with multiple proctors to generate high-precision labels to define $s$.

Specifically, given the $i$-th test session, $\vai_H=\pi_H(\vXi)$ is the labeling result from a proctor without using the gaze plot, $\vai_M=\pi_M(\vXi)$ is the labeling result made by thresholding the predicted gaze directions in each frame, and $\vai_{H+M}=\pi_{H+M}(\vXi)$ is the labeling result from a proctor using the gaze plot. For the three binary vectors $\vai_H$, $\vai_M$, and $\vai_{H+M}$, we collect all the positive intervals, and present the intervals for a group of $K$ proctors to label (without gaze plot), and take the majority opinion $\va^{*(i)}$ as the reference for comparison.

Note that we ensure high precision for $\va^{*(i)}$ by using multiple proctors to reduce variance and selecting only positive intervals instead of the entire video to reduce tediousness. However, this also means that if an interval is labeled as negative by all three systems, it will not be labeled differently in this process.

Comparing $\vai_H$, $\vai_M$, and $\vai_{H+M}$ with $\va^{*(i)}$, we can calculate the average precision and (upper-bounded) recall of each system as $F(\pi_{H+M})$, $F(\pi_{H})$, and $F(\pi_{M})$. Conducting this experiment is the next step in this project.

\section{User Study}

We also conducted a user study, where we recruited 11 DET proctors to try out the AI-assisted gaze detection system on 300 test sessions sampled from DET. The proctoring results were not used for the official certification, but we collected the feedback from the proctors regarding the gaze detection system with a survey form.

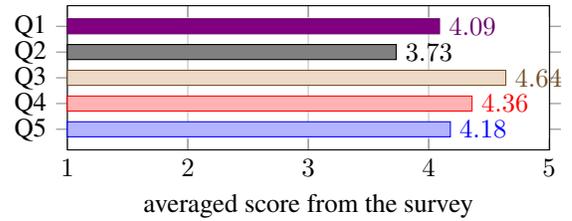
\begin{figure}[t]
\centering
\pgfplotstableread[header=false]{
4.18 Q5
4.36 Q4
4.64 Q3
3.73 Q2
4.09 Q1
}\datatable

\begin{tikzpicture}
  \begin{axis}[
        xbar, bar shift=0pt,
        enlarge y limits=0.2,
        xmin=1,
        ytick={0,...,4},
        yticklabels from table={\datatable}{1},
        xtick={1, 2, 3, 4, 5},
        xmajorgrids = true,
        bar width=2mm,
        width=0.95\columnwidth, height=3.5cm,
        xlabel={averaged score from the survey},
        nodes near coords, nodes near coords align={horizontal},
  ]

\pgfplotsinvokeforeach{0,...,4}{
    \addplot table [select row=#1, y expr=#1] {\datatable};
}
\end{axis}
\end{tikzpicture}
\caption{Survey questions\footnotemark: (Q1) I felt comfortable utilizing the tool, (Q2) I felt confident that the tool was providing me with correct information, (Q3) I felt the documentation/videos provided allowed me to easily understand how to use the tool, (Q4) I didn't have difficulty interpreting or understanding any visual elements of the tool, (Q5) I found it easy to incorporate the tool in my normal proctoring processes.}

\label{fig.survey}
\end{figure}

The survey is based on a scale of 1-5, where 1 represents ``absolutely disagree'' and 5 represents ``absolutely agree''. Here we show the survey questions and the final averaged scores in Figure~\ref{fig.survey}. Positive responses were received in the user study.

\footnotetext{Questions that are related to the details of the internal proctoring process are omitted, and the expression of Q4 and its score were reversed to make the interpretation of the scores more consistent. The original Q4 asked if proctors had difficulty.}

\section{Conclusion}

This paper presents an AI-assisted gaze detection system, which enables proctors to work effectively in finding the moments where a test taker is looking away from the screen. For the demo, we plan to show the gaze detection system on a laptop with an example test session, and the audience would be able to play with the system and give us feedback.

\subsection{Limitations}
We acknowledge that our system still has limitations, and future work will be needed to further improve the design. Firstly, the gaze plot only shows the gaze directions of the test takers, it doesn't show where on the screen the test taker is actually looking at. Therefore, the gaze plot should not directly be used alone to determine if the test taker is looking away. We expect the ML-only system to perform poorly because calibration will be needed to determine the exact relative positional relationships between the screen, the camera, and the test taker. Secondly, our system currently only works in an asynchronous proctoring environment, where the exam video is recorded. If synchronous proctoring is required, real-time prediction would be needed and the predictions need to be gradually added into the gaze plot. Finally, our proposed evaluation is still based on proctor decisions, and therefore is limited by the information that could be derived from the recorded information. To further improve the accuracy of the evaluation, we could have test takers taking the exams in a controlled environment where the camera and the screen are carefully calibrated. This will allow us to gather accurate measurements for test takers' eye gazes.

\section{Acknowledgments}
We thank the colleagues who had tested the system, provided feedback, reviewed the code, or reviewed the paper.

\bibliography{aaai24}

\end{document}